# A Wiki for Business Rules in Open Vocabulary, Executable English

**Adrian Walker**
Reengineering
Bristol, CT 06011-1412, USA
internet.business.logic@gmail.com

**Abstract**

The problem of business-IT alignment is of widespread economic concern.

As one way of addressing the problem, this paper describes an online system that functions as a kind of Wiki -- one that supports the collaborative writing and running of business and scientific applications, as rules in open vocabulary, executable English, using a browser.

Since the rules are in English, they are indexed by Google and other search engines.  This is useful when looking for rules for a task that one has in mind.

The design of the system integrates the semantics of data, with a semantics of an inference method, and also with the meanings of English sentences.  As such, the system has functionality that may be useful for the Rules, Logic, Proof and Trust requirements of the Semantic Web.

The system accepts rules, and small numbers of facts, typed or copy-pasted directly into a browser.  One can then run the rules, again using a browser.  For larger amounts of data, the system uses information in the rules to automatically generate and run SQL over networked databases.  From a few highly declarative rules, the system typically generates SQL that would be too complicated to write reliably by hand.  However, the system can explain its results in step-by-step hypertexted English, at the business or scientific level

As befits a Wiki, shared use of the system is free.

**Introduction**

The well known "layer cake" diagram (Berners-Lee 2004) outlines a high level agenda for work on the Semantic Web.

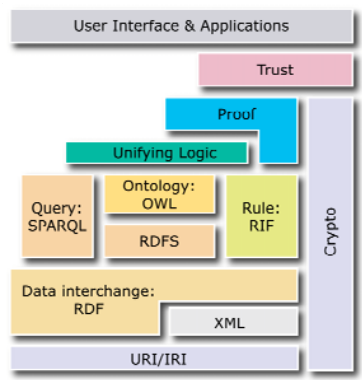

Figure 1. The Semantic Web Layer Cake

There is much current work in progress under the heading of "Semantics", as in the interleaving of metadata with data in RDF or OWL, stemming from (Berners-Lee et al, 2001). Such work fits into the layers from XML to Ontology in Figure 1. It may be useful to think of this as "data semantics", or "Semantics1".

In the diagram, there are boxes labeled Rule, Unifying Logic, Proof and Trust, User Interface and Applications. This paper describes one way of meeting some of the requirements indicated by those boxes, with an online system that combines Semantics1 with two further kinds of meaning.

Semantics2 specifies what conclusions a reasoning engine should be able to infer from any set of rules and facts (Walker 1993), using a logical model theory (Apt et al 1988) that takes into account the semantics under which databases are used in practice. Semantics3 concerns the meaning of English concepts at the author- and user-interface. The design of the system described here rests on making Semantics1, 2, and 3 work together.

Adding and integrating these semantic dimensions has the potential not only to support aspects of the Semantic Web, but also to ease some significant problems in commercial IT. According to (Forrester 2005) : *Aligning IT strategy with business strategy has been one of the top three issues confronting IT and business executives for more than 20 years. Polling of CIOs and business executives conducted in 2004 revealed that aligning IT and business goals remains their No. 1 or 2 priority*.

As one way of addressing the problem, this paper shows how to support the writing and running of applications as rules in open vocabulary, executable English. The system that does this takes a lightweight approach to English, backed by an inference engine that assigns a highly declarative meaning to a set of rules. The system accepts rules in English, and small numbers of RDF or relational facts, typed or copy-pasted directly into a browser. For larger amounts of data, the system uses information in the rules to automatically generate and run SQL over networked databases. From a few rules, the system typically generates SQL that would be too complicated to write reliably by hand. However, the system can explain its results in step-by-step hypertexted English, at the business or scientific level.

From the point of view of a person writing and running rules, the system is simply a browser, used actively rather than for passive browsing. Thus, the system functions as a kind of Wiki for executable English content. Here are some examples.

**A Semantic Resolution Example**

The paper (Peng et al 2002) describes an example of name resolution for e-commerce, using three namespaces: retailer, manufacturer, and shared.

In the example, a retailer orders computers from a manufacturer. However, in the retailer's terminology, a computer is called a PC for Gamers, while in the manufacturer's terminology, it is called a Prof Desktop.

Fortunately, the retailer and the manufacturer can agree that both belong to the class of Workstations/Desktops.  We also find out to what extent a Prof Desktop has the required memory, CPU and so forth for a PC for Gamers.

One of the rules for this is shown in Figure 2.

> for  the retailer  the term some-item1 has super-class some-class in the some-ns namespace
> for  the manufacturer  the term some-item2 has super-class that-class in the that-ns namespace
> ---------------------------------------------------------------------------------------------------------------------------------
> the retailer term that-item1 and the manufacturer term that-item2 agree - they are of type that-class

Figure 2.  A Rule for Semantic Resolution

The rule says that, if the two premises above the line are true, then so is the conclusion.  In the rule, "some-item1", "that-class" and so on are place holders, or variables, that will be filled in with actual values, such as "PC for Gamers" and "Computers" when the rule is run.  Apart from the place holders, the rest of the words in the rule are in open vocabulary, open syntax, English.  So, the rule defines the meaning of the last sentence in terms of the meanings of the first two.  To avoid infinite regress, this process stops at the headings of data tables.  The headings are similar sentences, and the number of place holders in a heading is the number of columns in the table.

To view and run the example, one can point a browser to (Reeng 2006) and select SemanticResolution1.

**An RDF Query Example**

The paper (Haase et al 2004) describes a use case in which 14 test questions are put to some RDF data about published papers and their authors.  The paper describes several Semantic Web query languages, and shows that none of those languages can answer all 14 questions.

It is straightforward to write rules that allow the system to answer all 14 questions.  One such rule is shown in Figure 3.

> some-paper is related by fact#:title to some-title
> that-paper is related by fact#:author to some-description
> that-description is related by some-rdf-node to some-home-page
> that-home-page is related by fact#:name to some-name
> that-home-page is related by fact#:email to some-email
> ------------------------------------------------------------------------------------
> that-name is an author , with email that-email , of that-title

Figure 3.  A Rule for the RDF Query Example

Reasoning over RDF, even in this relatively simple example, is quite hard for a person to follow. A nontechnical user, who was unsure whether to trust an answer, would have a hard time convincing himself of its validity simply by looking at the rules and the RDF data, and would probably find it impossible to do so with a SQL-like query language. To help with this matter of trust, the system can supply a step-by-step English explanation of any answer that it produces. It can also explain in English why it failed to give an expected answer.

The answer to a request such as "show me the authors of papers and their email addresses" is a table saying that, amongst others, "Jeen Broekstra is an author , with email jbroeks@cs.vu.nl , of RDF Query Languages". A step in an explanation is shown in Figure 4.

```
Paper is related by fact#:title to  An Overview of RDF Query Languages
Paper is related by fact#:author to __Description1
__Description1 is related by rdf:_1 to http://www.cs.vu.nl/~jbroeks/
http://www.cs.vu.nl/~jbroeks/ is related by fact#:name to  Jeen Broekstra
http://www.cs.vu.nl/~jbroeks/ is related by fact#:email to jbroeks@cs.vu.nl
-----------------------------------------------------------------------------------------------------
Jeen Broekstra  is an author , with email jbroeks@cs.vu.nl , of  RDF Query Languages
```

Figure 4. An Explanation Step in the RDF Query Example

On the other hand, if we ask whether Adrian Walker is the author of the paper, we get a "No" answer, and a step in the explanation is similar to Figure 4, except that the last two premises are marked as "missing", and the conclusion is marked "not shown".

An explanation always starts out with the general justification of an answer, and provides hyperlinks so that one can drill down into more detail as needed. In particular, we could write additional rules so that a rather technical, RDF-based explanation step is preceded by something more suitable for an end user to read.

To view and run the example, one can point a browser to (Reeng 2006) and select RDFQueryLangComparison1.

**An OWL Inferencing Test Example**

The W3C provides a number of test cases for OWL (W3C 2004). One of these requires the inference that the items in a list are different if they are of rdf:type owl:AllDifferent.

One of the rules for this task is shown in Figure 5.

```
some-tag is related by rdf:type to owl:AllDifferent
that-tag is related by owl:distinctMembers to some-start-tag
from that-start-tag we can follow a list to some-item
that-item is related by rdf:type to some-type
-----------------------------------------------------------------------------------------------------
that-tag names a collection of distinct items of type that-type that includes that-item
```

Figure 5.   A Rule for the OWL Inferencing Example

To view and run the example, one can point a browser to (Reeng 2006) and select OwlTest1.

**An Oil Industry Supply Chain Example**

When a geographic region has a demand for a quantity of an oil product, it is in general possible to meet the demand using a number of equivalent products. Many factors influence the proportions of component products that are combined to make an optimal supply chain decision. The factors include the season of the year, the locations of available equivalent products, and the availability of suitable and timely transportation.

For our example (Kowalski and Walker 2005), we project that the target region NJ will need 1000 gallons of product 'y' in October. We then ask what alternative routes and modes-of-transportation (truck, train, boat, pipe) do we have to get that product to the region. Next we ask whether there's a refinery nearby that can produce the base product for finished product 'y'. With all of that, we finally say that we need a delivery plan that is optimized to deliver on time, make a profit, and beat the competition.  However, if there is not enough of product 'y', then, depending on the region and the customers, product 'x' or 'z' will do as well; they're just variations of 'y' using different additives. But they'll only do just as well in region NJ for the season including October. This makes sales projections and marketing more complicated, but also gives us more competitive flexibility.

One of the rules for the supply chain task is shown in Figure 6.

```
Estimated demand some-id in some-region is for some-quantity gallons of some-finished-product
    in some-month of some-year
for demand that-id for that-finished-product refinery some-refinery can supply some-amount gallons of some-product
for demand that-id the refineries have altogether some-total gallons of acceptable base products
that-amount / that-total = some-long-fraction
that-long-fraction rounded to 2 places after the decimal point is some-fraction
-------------------------------------------------------------------------------------------------------------------------------
for estimated demand that-id that-fraction of the order will be that-product from that-refinery
```

Figure 6.  A Rule for the Oil Industry Supply Chain Example

 As of the writing of this paper, typing "estimated demand some-id in some-region" into Google finds the rules on the world wide web.

Here is a fragment of the SQL that is automatically generated from the rules:

*select distinct x6,T2.PRODUCT,T1.NAME,T2.AMOUNT,x5 from*
*T6 tt1,T6 tt2,T5,T4,T3,T2,T1,T6,*
*(select x3 x6,T6.FINISHED_PRODUCT x7,T6.ID x8,tt1.ID x9,tt2.ID x10,sum(x4) x5 from*
*T6,T6 tt1,T6 tt2,*
*((select T6.ID x3,T3.PRODUCT1,T1.NAME,T2.AMOUNT x4,T2.PRODUCT from*
*T1,T2,T3,T4,T5,T6,T6 tt1,T6 tt2 where*
*T1.NAME=T2.NAME and T1.REGION=T6.REGION and T2.MONTH1=T4.MONTH1 and*
*T2.MONTH1=T6.MONTH1 and T2.PRODUCT=T3.PRODUCT2 and T4.MONTH1=T6.MONTH1 and*
*T3.PRODUCT1=T6.FINISHED_PRODUCT and T3.SEASON=T4.SEASON and*
*T3.SEASON=T5.SEASON and*
*T4.SEASON=T5.SEASON and T6.ID=tt1.ID and T6.ID=tt2.ID and tt1.ID=tt2.ID)*
*union ....*

It would be difficult to write such SQL reliably by hand, or to manually validate a result that the system has found. As a way of establishing trust, the system can explain each result in step-by-step, hypertexted English, at the business or scientific level. To view and run the example, one can point a browser to (Reeng 2006) and select Oil-IndustrySupplyChain1.

**A Bioinformatics Ontology Example**

The paper (Smith et al 2005) describes a way of assigning a formal meaning to relations in bioinformatics ontologies. For example, the paper defines what it means for a continuant class to be a part of another class, in terms of some easily understood primitive relations and some reasoning over time.

If C and C1 are classes, then the paper defines

*C part_of C1 = [definition] for all c, t, if Cct then*
 *there is some c1 such that C1c1t and c part_of c1 at t*

where *Cct* is shorthand for "c is an instance of C at time t"

Figure 7 shows some of the rules that make this definition executable over ground data.

```
for all c, t, if some-C c t then there is some c1 such that some-C1 c1 t and c part_of c1 at t
-----------------------------------------------------------------------------------------------
that-C is a part_of the continuant class that-C1

(A c,t) [ some-C c t => (E c1) [ some-C1 c1 t and c part_of c1 at t ] ]
-----------------------------------------------------------------------------------------------
for all c, t, if that-C c t then there is some c1 such that that-C1 c1 t and c part_of c1 at t

some-C and some-C1 are two different Non-process classes with instances
not : (E c,t) [ that-C c t and not (E c1) [ that-C1 c1 t and c part_of c1 at t ] ]
-----------------------------------------------------------------------------------------
(A c,t) [ that-C c t => (E c1) [ that-C1 c1 t and c part_of c1 at t ] ]
```

Figure 7. Some Rules for the Bioinformatics Ontology Example

To view and run the example, one can point a browser to (Reeng 2006) and select RelBioOntDefn3.

**System Design**

In any system designed to answer English questions, we need a way of showing users what kinds of questions can reasonably be asked. If we load some oil industry rules into the system, we do not want a user to ask about bioinformatics ontologies. The current system guides a user by showing a menu consisting of generalized versions of the sentences that were used in writing the rules for a particular task. At the top of the menu, the system shows a group of sentences that are not used as premises in any rules. These correspond to generalized versions of the most important questions that the application can answer. Next, the system shows sentences that were used as premises for the top group, and so on.

A user can select a sentence directly from the menu. Alternatively, she can type in an arbitrary English sentence, in which case the system will use a form of conventional information retrieval to bring relevant rule conclusions to the top of the menu.

Once a user has selected a generalized sentence to use as a question, she can specialize it using automatically generated submenus. This can be done by replacing "some-name" with "Fred", or making a range restriction, or asking for an approximate match.

For a person writing rules, the vocabulary is *open*, and so, to a large extent, is the English syntax. In contrast to most natural language systems, there is no dictionary or grammar of English to be maintained. As such, the support for English is lightweight, but this has a practical advantage -- someone writing rules can freely use jargon, acronyms, words not in any dictionary, ungrammatical sentences, mathematical notations and so on.  The price for this is, if an author wishes the system to regard two sentences as the same, she must write rules that say so.  An advantage is that the English semantics are strict – a sentence means exactly what it says (in context), and nothing else.  Although the vocabulary is open as far as the system is concerned, one can write a collection of rules and facts that defines a controlled vocabulary.

While there is no dictionary or grammar of English, the system does contain a grammar that recognizes rules and tables of data.  So, if an author types in a simple rule that does not consist of one or more sentences, followed by an underline, followed by a single sentence, the system responds with a warning.  Also, the underlying rule syntax requires that each place holder (or variable) that appears in a conclusion of a rule should also appear in simple premise of the same rule.  Similarly, if a sentence is negated (as in the Bioinformatics Ontology Example), the system requires that each of the variables in the sentence should be mentioned in earlier simple premises.  There is further checking to ensure that it is not possible for the conclusion of a rule to depend on a negation of itself, even if that happens through a chain of other rules.  Should that happen, the system shows the rules concerned, together with a warning.

The lightweight English -- Semantics3 -- is supported by an inference method that assigns a strongly non-procedural meaning to the rules -- Semantics2 -- (Apt et al 1988).

The inference method used is based on that in (Walker 1993).  The method switches automatically between forward and backward execution of the rules.

The non-procedural approach is necessary, because an English sentence must keep the same declarative meaning regardless of where it appears sequentially in a collection of rules, and regardless of changes to rules in which the sentence does not appear.  For authors who are not programmers, this approach has the advantage that the answers one gets from a set of rules and facts do not depend on the textual sequence in which the rules appear.  This is in contrast to a procedural reading, in which a collection of N rules potentially has factorial(N) different meanings, corresponding to different permutations of the rule sequence.

**Conclusions**

I have described a design that combines semantics for data, for inference, and for English.  There is an online system for this -- a kind of Wiki for rules in open vocabulary, executable English.  When the rules are run, the system can explain its results (and also why expected results are missing), in hypertexted English at the business or scientific level.  The system can also automatically generate and run SQL over networked databases.  The generated SQL can be too complex to write reliably by hand, but its results can also be explained in English.  The author- and user-interface of the system is simply a browser, and shared use of the system is free.  Since the rules are in English, they are indexed by Google and other search engines.  This is useful when looking for rules for a task that one has in mind.

The approach used in the system has the potential to address some of the requirements for Rules, Logic, Proof and Trust that are outlined in the upper levels of the Semantic Web layer cake diagram: Rules and Logic via highly declarative inferencing; Proof and Trust via English explanations. Similarly, there is the potential for the approach to address the significant economic problem of business-IT alignment.